\documentclass[10pt,letterpaper]{article}

\usepackage[margin=1in]{geometry}
\usepackage[T1]{fontenc}
\usepackage[utf8]{inputenc}
\usepackage{amsmath}
\usepackage{amssymb}
\usepackage{amsthm}               
\usepackage{newtxtext}            
\usepackage[varvw]{newtxmath}     
\usepackage[protrusion=false]{microtype} 
\usepackage{graphicx}
\usepackage{booktabs}
\usepackage{xcolor}
\usepackage[font=normalsize,labelfont=bf,labelsep=period]{caption}
\usepackage{xurl}                 
\usepackage[hidelinks]{hyperref}  
\usepackage{orcidlink}            


\title{Agentic AI uncovers conserved cross-tissue protein co-abundance programs inaccessible to single-dataset analysis}
\author{%
  Dehao Wu\textsuperscript{1,$\dagger$}\,\orcidlink{0009-0008-8209-267X},\quad
  Runyu Guan\textsuperscript{1,$\dagger$}\,\orcidlink{0009-0006-1890-3349},\quad
  Qiqi Xie\textsuperscript{1}\,\orcidlink{0000-0003-4099-5287},\\
  Yang Li\textsuperscript{2}\,\orcidlink{0000-0003-2530-7167},\quad
  Haohan Wang\textsuperscript{1,$\ast$}\,\orcidlink{0000-0002-1826-4069}\\[1.5ex]
  \normalsize
  \textsuperscript{1}School of Information Sciences, University of Illinois Urbana-Champaign, Champaign, IL 61820, US\\
  \normalsize
  \textsuperscript{2}Department of Ecology and Evolutionary Biology, University of Michigan, Ann Arbor, MI 48109, US\\[1ex]
  \normalsize
  \textsuperscript{$\dagger$}These authors contributed equally to this work.\\
  \normalsize
  \textsuperscript{$\ast$}Corresponding author:
  \href{mailto:haohanw@illinois.edu}{\texttt{haohanw@illinois.edu}}
}
\date{}

\begin{document}

\maketitle

\begin{abstract}
\noindent
Protein co-abundance clusters preserved across tissues can nominate shared disease mechanisms and candidate therapeutic targets, particularly where proteins implicated in organ-confined diseases converge in other tissues. Yet cross-tissue proteomic studies have examined biologically pre-selected tissue pairs, leaving the majority of possible combinations, and the non-obvious relationships among them, unexplored. Exhaustive comparison could close this gap but requires assessing thousands of candidate co-abundance clusters from hundreds of tissue-pair networks consistently across heterogeneous evidence types. This scale exceeds what manual curation can cover. We therefore present an LLM-agent framework for large-scale, evidence-grounded comparison of tissue-specific protein co-abundance networks. Starting from sample-level proteomic data, the framework constructs tissue networks, derives conserved co-abundance clusters for each tissue pair, integrates evidence from expression atlases, protein interaction and complex databases, pathway annotations, disease associations and the literature, and proposes mechanistic hypotheses in which every conclusion cites its evidence. Applied to all 231 pairs among 22 human tissues and sample groups, the framework identified 1{,}825 conserved co-abundance clusters in 216 tissue pairs. Co-abundance relationships were shared beyond chance in 76\% of tissue pairs, including anatomically unrelated organs, and which clusters two tissues share was only weakly associated with their abundance similarity. The clusters were enriched 3.7- to 6.9-fold for known interactions and complex memberships, and 379 of 793 testable clusters replicated in independent tissue proteomes from CPTAC. Beyond recovering known biology, some clusters pointed to relationships not recorded in current databases: a cluster shared by liver and bone marrow brought together three genes that cause inherited white matter disorders, suggesting a common mechanism through the integrated stress response. Together, these results indicate that conserved co-abundance clusters point to biological processes shared between tissues, providing a hypothesis-generating resource for studying mechanisms and disease-associated proteins beyond the affected organ. Code and data are available at \url{https://github.com/jacob20030316-crypto/Protein_Co_Abundance_Agent}.

\end{abstract}

\section{Introduction}

Cross-tissue protein coordination contributes to disease across organ systems, and several inter-organ axes have been traced from molecular mechanism to clinical intervention. Liver-secreted FGF21 signals through the central nervous system to suppress hepatic \textit{de novo} lipogenesis, and FGF21 analogues are in late-stage trials for MASH~\cite{rose2025fgf21}. Liver-derived hepcidin restricts iron availability to bone marrow and intestine, and the hepcidin mimetic rusfertide reduced phlebotomy dependence in polycythemia vera~\cite{kremyanskaya2024rusfertide}. Proteomic comparison across compartments is also generating candidates where such axes are less mature. Plasma protein programs, including the matrisome-associated protein SMOC1, track cerebral amyloidosis and cognitive decline, yet roughly half of cognition-associated plasma proteins remain unexplained by measured brain neuropathology~\cite{afshar2025plasma}. In amyotrophic lateral sclerosis, plasma protein changes are concentrated in skeletal muscle and energy metabolism rather than in the motor neurons where the disease begins~\cite{chia2025als}. Disease-associated protein changes therefore need not be confined to the affected organ.

Whether such changes reflect protein co-abundance clusters shared across tissues has rarely been examined systematically. At the transcript level, co-expression networks have been compared across GTEx tissues, and some co-expression modules recur in several tissues~\cite{pierson2015gnat,russell2023communities}. At the protein level, co-abundance maps have pooled samples from many conditions into a single network~\cite{kustatscher2019proteomehd}, leaving open which co-abundance relationships each pair of tissues shares. Cross-tissue proteomic studies such as those above compare compartments chosen from prior biological knowledge~\cite{afshar2025plasma,chia2025als}, and combinations that no specific hypothesis has motivated have seldom been studied together. Exhaustive comparison across all tissue pairs could nominate shared disease mechanisms and candidate therapeutic targets in these overlooked combinations. Such comparison would be particularly informative where proteins implicated in organ-confined diseases converge within co-abundance clusters shared with other tissues, suggesting entry points for study and intervention outside the primary site of pathology. Yet the analysis it requires is extensive, as heterogeneous tissue data must first be checked and matched across resources and hundreds of tissue pairs each yield several co-abundance clusters. Every cluster must then be interpreted against expression atlases, protein--protein interactions, complex annotations, pathway membership, disease associations and the primary literature. The scale of this analysis is beyond manual curation.

\begin{figure}[tbp]
  \centering
  \includegraphics[width=0.95\textwidth]{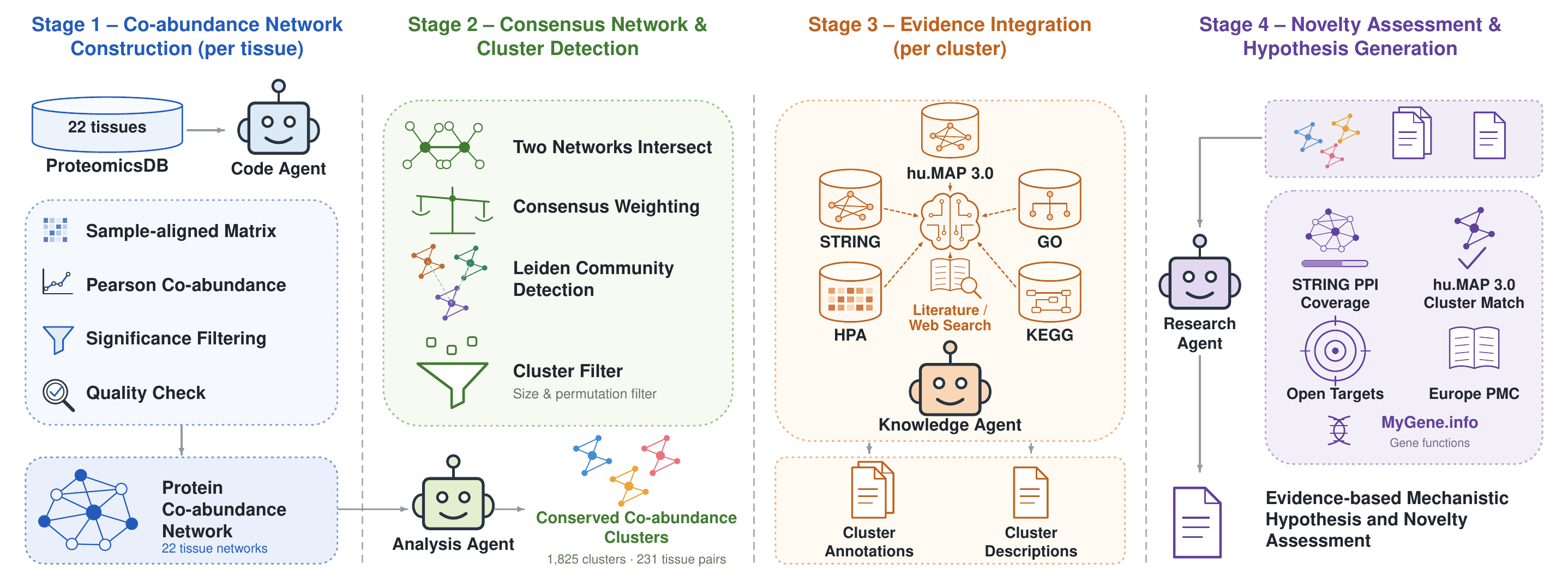}
  \caption{Overview of the LLM-agent framework. Stage~1, co-abundance network construction for each tissue; Stage~2, consensus networks and conserved co-abundance clusters for each tissue pair; Stage~3, evidence integration for each cluster; Stage~4, hypothesis generation and novelty assessment.}
  \label{fig:pipeline}
\end{figure}

We therefore developed an LLM-agent framework for systematic cross-tissue comparison of human protein co-abundance (Figure~\ref{fig:pipeline}). Starting from sample-level proteomic data in ProteomicsDB, the agents autonomously build a co-abundance network for each tissue and identify conserved co-abundance clusters between tissue pairs. For every cluster, they retrieve evidence from protein atlases, interaction and complex databases, pathways, disease associations and the literature. On this basis, they infer the function of each cluster, assess its novelty and propose mechanistic hypotheses, citing the evidence behind every conclusion. The agents carry out every step from data retrieval to interpretation, so the analysis extends to every tissue pair. Because all evidence is retrieved through database tools and cited in each conclusion, every interpretation remains traceable at this scale. With this framework, we compared all 231 pairs among 22 tissues and sample groups, identified 1{,}825 conserved co-abundance clusters and produced an evidence-supported interpretation for each.

The comparison showed that protein co-abundance relationships were widely shared between tissues, often between organs with no direct anatomical or functional connection, such as colon and lung. Which co-abundance clusters two tissues share was only weakly associated with their protein abundance similarity: brain--lung and autonomic nervous system--bone had nearly identical abundance similarity, yet shared 20 and 2 conserved co-abundance clusters. Choosing tissue pairs in advance, for example by their similarity, would therefore miss many of these relationships. Many shared clusters recovered known biology and recurred in independent cohorts; prostate and kidney, for example, shared a cluster spanning the branched-chain amino acid degradation pathway that was also co-abundant in an independent kidney cohort. Others pointed to relationships not yet described: a cluster shared by liver and bone marrow brought together three genes that cause inherited white matter disorders, suggesting a common mechanism.

Our contributions are as follows:
\begin{itemize}
  \item An LLM-agent framework that carries out exhaustive cross-tissue comparison end to end, from sample-level proteomic data to hypotheses in which every conclusion is traceable to retrieved evidence.
  \item An atlas of conserved co-abundance clusters across all 231 tissue pairs, showing where each organ's clusters recur and thus where disease-relevant clusters can be studied beyond the affected organ.
  \item Evidence-cited mechanistic hypotheses for every cluster, providing candidates for experimental testing.
\end{itemize}

\section{Methods}
\label{sec:methods}

\subsection{Agent framework}
\label{sec:agents}

The framework takes sample-level proteomes from ProteomicsDB as input. Of the 45 tissues and body fluids in ProteomicsDB, we retained the 22 tissues and sample groups with enough samples to build a co-abundance network, which form 231 tissue pairs. Four agent roles divide the workflow from these proteomes to interpreted clusters, each responsible for one stage in Figure~\ref{fig:pipeline} and equipped with the tools that its stage requires.
\begin{itemize}
  \item The \emph{Code Agent} reads the sample-aligned abundance matrix of each tissue, builds its co-abundance network and checks the quality of the network.
  \item The \emph{Analysis Agent} builds the consensus network of each tissue pair and identifies its conserved co-abundance clusters.
  \item The \emph{Knowledge Agent} retrieves protein atlas, interaction, complex, pathway enrichment and literature evidence for each cluster and writes its annotation and description.
  \item The \emph{Research Agent} reads all evidence for a cluster, together with the function, disease associations and publication counts of its member genes, and then infers the function of the cluster, assesses its novelty and proposes mechanistic hypotheses.
\end{itemize}
The output of each stage forms the input of the next. The agents call tools by writing and executing code. To keep every conclusion traceable, all evidence is retrieved through tools, the agent cites the evidence entries behind each conclusion, and the values in a report are taken directly from the retrieved records. When an output lacks required content or cites evidence that was not retrieved, the specific problem is returned to the agent, which revises and resubmits the output. The Code Agent and the Analysis Agent use \texttt{gpt-4.1}, and the Knowledge Agent and the Research Agent use \texttt{gpt-5.6-sol}.

\subsection{From tissue proteomes to conserved co-abundance clusters}
\label{sec:clusters}

The Code Agent and the Analysis Agent derive the conserved co-abundance clusters from tissue proteomes in ProteomicsDB~\cite{proteomicsdb2018}, where samples of the same tissue come from different studies and measure different sets of proteins. Abundance values (iBAQ) were therefore retrieved sample by sample and aligned by sample identifier into a sample-by-protein matrix for each tissue, with proteins identified by UniProt~\cite{uniprot2025}.

The network tool computes the correlation between two proteins from the samples in which both were measured. For proteins $i$ and $j$ in tissue $t$, let $S^{(t)}_{ij}$ be this set of samples and $r^{(t)}_{ij}$ the Pearson correlation computed on it. $P$ values were computed from the $t$ statistic with $|S^{(t)}_{ij}| - 2$ degrees of freedom, which accounts for the varying number of shared samples, and adjusted across all protein pairs within the tissue with the Benjamini--Hochberg procedure~\cite{benjamini1995controlling}, giving $q^{(t)}_{ij}$. The tool links two proteins only when their correlation is supported by enough samples, statistically significant and strong:
\begin{equation}
  E_t = \left\{ (i,j) : |S^{(t)}_{ij}| \ge 10,\; q^{(t)}_{ij} < 0.05,\; |r^{(t)}_{ij}| \ge 0.8 \right\},
  \label{eq:edges}
\end{equation}
with edge weights $w^{(t)}_{ij} = r^{(t)}_{ij}$. The same rule is applied to every tissue and yielded a co-abundance network for each of the 22 tissues and sample groups.

Conserved co-abundance must be present in both tissues with the same sign. The consensus network of a tissue pair $(a,b)$ therefore keeps only edges that are present in both networks and positive in both, weighted by the weaker of the two:
\begin{equation}
  w^{(ab)}_{ij} = \min\left(w^{(a)}_{ij},\, w^{(b)}_{ij}\right) > 0, \qquad (i,j) \in E_a \cap E_b .
  \label{eq:consensus}
\end{equation}
Its nodes are the proteins measured in both tissues, excluding housekeeping genes and proteins without a gene annotation, so that ubiquitously expressed proteins do not dominate the network structure.

Groups of proteins that are co-abundant in both tissues appear as densely connected communities of the consensus network. The clustering tool identifies them with the Leiden algorithm~\cite{traag2019leiden}, maximizing weighted modularity at resolution $\gamma = 1$. Communities of 5 to 50 proteins were retained as candidate clusters. To exclude communities that arise only because highly connected proteins group together by chance, protein labels in one tissue were permuted 10{,}000 times among proteins of similar degree (strata of 20 proteins), with each tissue held fixed in turn. Candidates whose shared within-cluster edges did not exceed the permutation distribution (Benjamini--Hochberg $q < 0.05$) were removed. The remaining 1{,}825 clusters are the conserved co-abundance clusters analyzed in this study.

\subsection{Evidence integration and hypothesis generation}
\label{sec:evidence}

The Knowledge Agent assembles the evidence for each cluster, starting with gene annotations from the Human Protein Atlas~\cite{uhlen2015hpa}, including immunohistochemistry in the corresponding normal tissues. It also retrieves protein complexes in hu.MAP~3.0 containing at least two cluster members~\cite{fischer2025humap3}, STRING interactions with a combined score of at least 700 and STRING functional enrichment~\cite{szklarczyk2023string}, GO~\cite{go2000,go2026} and KEGG~\cite{kanehisa2000kegg} enrichment, and web search results (DuckDuckGo). Because ProteomicsDB and the Human Protein Atlas group tissues differently, the agent decides for each tissue group which atlas tissues correspond to it.

GO and KEGG enrichment was assessed with a one-sided hypergeometric test, using the proteins of the tissue pair's consensus network as background, since only these proteins can enter a cluster. $P$ values were adjusted with the Benjamini--Hochberg procedure both within each cluster, to rank terms for that cluster, and across all cluster--term tests; the $q$ values reported in this paper are the latter. The Knowledge Agent integrates this evidence into an annotation and a description of each cluster.

Building on this evidence, the Research Agent interprets each cluster. It reads the evidence of the cluster together with the functional annotation of each member gene from MyGene.info~\cite{xin2016mygene}, its disease associations in Open Targets~\cite{ochoa2021opentargets} and its publication count in Europe PMC~\cite{rosonovski2024europepmc}. It then infers the function of the cluster, examines alternative explanations such as cell composition or blood-derived proteins, assesses novelty from publication counts, identifies key genes and proposes mechanistic hypotheses, citing the evidence behind each conclusion.

\section{Results and Analysis}

\subsection{Widespread sharing of protein co-abundance across tissues}
\label{sec:edge-sharing}

To test whether protein co-abundance relationships are specific to each tissue or shared between tissues, we compared the networks of every tissue pair. For each of the 231 pairs, we counted the positive co-abundance edges present in both networks (Section~\ref{sec:clusters}). Chance overlap was estimated from degree-preserving random networks, in which each network was rewired independently by swapping edge endpoints (ten attempts per edge) so that every protein kept its number of connections. Over 10{,}000 randomizations, we computed empirical $P$ values from the edges shared by the randomized networks and adjusted them across all 231 pairs with the Benjamini--Hochberg procedure.

\begin{figure}[tbp]
  \centering
  \includegraphics[width=0.55\textwidth]{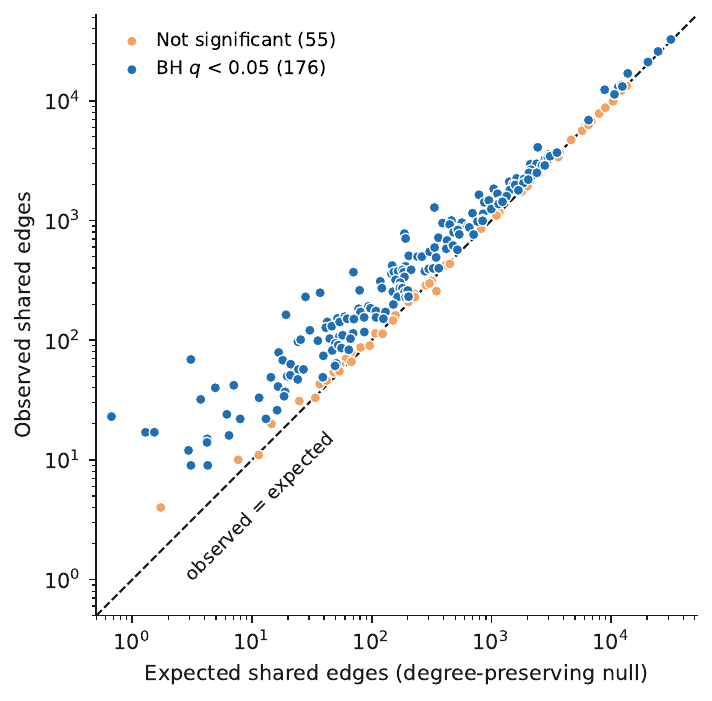}
  \caption{Shared co-abundance edges in all 231 tissue pairs. Each point is
  one pair; the dashed line marks equal observed and expected counts.}
  \label{fig:edge-sharing}
\end{figure}

In 176 of the 231 pairs (76.2\%), the two networks shared significantly more edges than expected by chance ($q < 0.05$; Figure~\ref{fig:edge-sharing}). Nearly all pairs lie on or above the diagonal, and 66 exceeded twice their expected overlap. Colon and rectum, adjacent intestinal segments, shared 163 edges against 19 expected (8.6-fold). Comparably strong sharing linked colon and lung (8.2-fold), organs with no direct anatomical or functional connection. Substantial sharing was also observed between brain and ovary (6.7-fold) and between kidney and myometrium (4.0-fold). Co-abundance relationships are therefore shared between most tissue pairs, including anatomically unrelated organs.

\subsection{An atlas of tissue relationships based on shared co-abundance clusters}
\label{sec:atlas}

\begin{table}[tbp]
  \centering
  \caption{Number of tissues or sample groups sharing at least one conserved
  protein co-abundance cluster with the reference tissue}
  \label{tab:shared-tissues}
  \begin{tabular}{lclc}
    \toprule
    Tissue or sample group & Count & Tissue or sample group & Count \\
    \midrule
    Bone & 21 & Liver & 20 \\
    Bone marrow & 21 & Prostate & 20 \\
    Brain & 21 & Skin & 20 \\
    Kidney & 21 & Stem cells & 20 \\
    Lung & 21 & Cerebrospinal fluid & 19 \\
    Rectum & 21 & Heart & 19 \\
    Testis & 21 & Myometrium & 18 \\
    Uterine cervix & 21 & Ovary & 18 \\
    Autonomic nervous system & 20 & Cell line & 17 \\
    Circulatory system & 20 & Placenta & 17 \\
    Colon & 20 & Breast & 16 \\
    \bottomrule
  \end{tabular}
\end{table}

Community detection on the consensus networks identified 1{,}825 conserved co-abundance clusters in 216 of the 231 tissue pairs (93.5\%), with a median of 7 per pair and at most 25 (Section~\ref{sec:clusters}). These clusters were not concentrated in a few tissues. Table~\ref{tab:shared-tissues} quantifies the breadth of sharing of each tissue or sample group as the number of other tissues sharing at least one conserved co-abundance cluster with it. Eight tissues or sample groups shared clusters with all 21 others, and breast, which shared clusters with the fewest tissues, still shared them with 16.

\begin{figure}[tbp]
  \centering
  \includegraphics[width=0.64\textwidth]{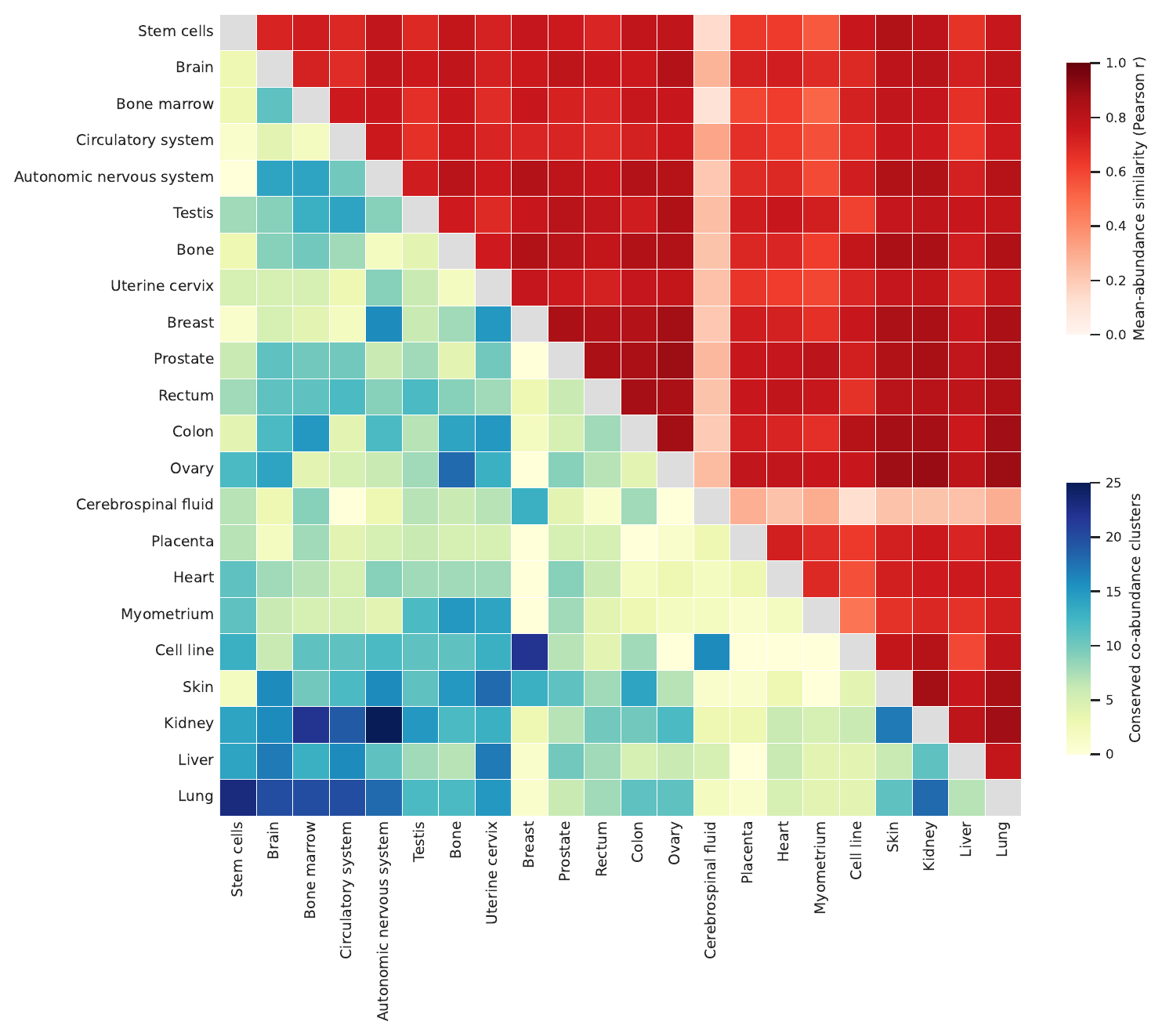}
  \caption{Pairwise comparison of 22 tissues and sample groups. Lower
  triangle, number of shared conserved co-abundance clusters; upper triangle,
  mean-abundance similarity (Pearson $r$).}
  \label{fig:atlas}
\end{figure}

Figure~\ref{fig:atlas} contrasts two views of tissue similarity. The lower triangle shows the number of shared conserved co-abundance clusters, and the upper triangle shows mean-abundance similarity, measured as the Pearson correlation of mean $\log_{10}$ abundance across proteins quantified in both tissues. Cluster sharing was structured: hierarchical clustering separated groups of tissues that share extensively from pairs that share few or none, whereas mean-abundance similarity was high for most pairs (median $r = 0.75$). Across the 231 pairs, the two measures were only weakly correlated (Spearman $\rho = 0.27$). Brain--lung and autonomic nervous system--bone, for example, had nearly identical mean-abundance correlations (0.794 and 0.806), yet the former shared 20 conserved co-abundance clusters and the latter only 2. Tissues therefore share not only isolated protein relationships but groups of co-abundant proteins, and which tissues share them is only weakly associated with how similar their mean abundances are.

\begin{figure}[t]
  \centering
  \includegraphics[width=0.95\textwidth]{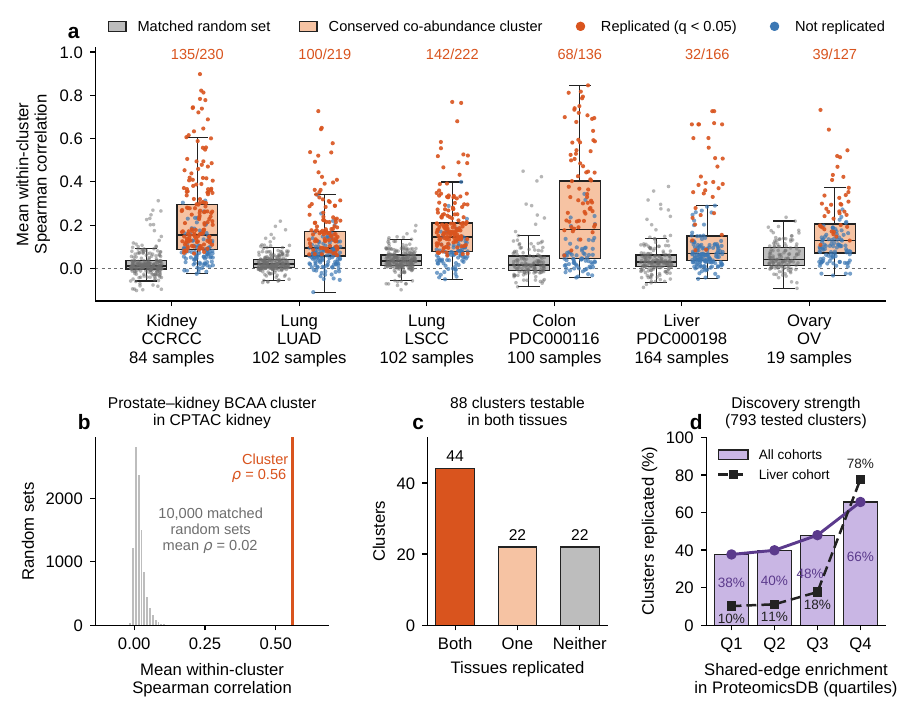}
  \caption{Replication of conserved co-abundance clusters in independent CPTAC normal tissue cohorts. (a) Mean within-cluster Spearman correlation of each testable cluster (orange box) and of one degree-matched random protein set drawn for each cluster (grey box). Orange points, replicated clusters (permutation $q < 0.05$, 10{,}000 matched random sets); blue points, clusters that did not replicate; numbers, replicated over testable clusters. (b) The branched-chain amino acid degradation cluster shared by prostate and kidney against its 10{,}000 matched random sets in the CPTAC kidney cohort. (c) Clusters testable in both of their tissues, by the number of tissues in which they replicated. (d) Replication by discovery strength, measured as the shared within-cluster edges in ProteomicsDB over their permutation expectation, in quartiles (Q1, 1.7- to 4.8-fold; Q4, 13- to 185-fold). The dashed line shows the liver cohort, which had the lowest overall replication, on the same quartile cut-offs.}
  \label{fig:replication}
\end{figure}

\subsection{Known interactions and replication in independent tissue proteomes}
\label{sec:replication}

The conserved co-abundance clusters were strongly enriched for known protein relationships. Compared with random protein sets matched for network degree, protein pairs within them were 5.0 times more often high-confidence STRING interactions (combined score $\geq 700$). They were also 3.7 and 6.9 times more often co-complex members in hu.MAP~3.0 and CORUM ($P \leq 10^{-4}$ for all three, 10{,}000 draws). This enrichment was not driven by a few large complexes. The 10\% of clusters with the most known pairs held 55\%, 53\% and 70\% of the known pairs in the three databases, yet the remaining clusters were still enriched 3.0-, 2.4- and 2.8-fold. As a set, the clusters therefore reflect biologically coherent relationships rather than chance co-variation.

To assess whether the clusters, identified in ProteomicsDB alone, reflect batch, laboratory or processing effects specific to this resource, we tested them in tumor-adjacent normal tissue proteomes from the Clinical Proteomic Tumor Analysis Consortium (CPTAC), covering kidney, lung, ovary, colon and liver in 571 samples~\cite{clark2019ccrcc,gillette2020luad,satpathy2021lscc,mcdermott2020ov,vasaikar2019coad,gao2019hcc}. These samples came from different individuals and laboratories and were generated with different quantification workflows.

Cluster membership was kept fixed, and no clusters were searched for in the independent data. For each cluster whose tissue pair included one of these tissues, we computed the mean pairwise Spearman correlation among its members in the matching cohort; a rank-based measure was used because the two resources quantify proteins on different scales. Only clusters with at least five members quantified in at least 80\% of samples were tested. As a reference, we drew random protein sets of the same size from the tissue pair's consensus network, matching each member on network degree (10{,}000 draws). This matching prevents highly connected proteins, which correlate more readily, from driving the comparison. The empirical $P$ value was the fraction of random sets reaching the observed value; after Benjamini--Hochberg adjustment within each cohort, clusters with $q < 0.05$ were considered replicated.

In every cohort, within-cluster correlation exceeded that of matched random sets (Figure~\ref{fig:replication}a). Of the 793 testable clusters, 379 replicated in at least one tissue. The most direct test came from clusters whose two tissues both had independent data. Of the 88 clusters that met this criterion in both of their tissues, 44 replicated in both (Figure~\ref{fig:replication}c), providing independent support for sharing between tissues.

Replication followed the strength of the evidence at discovery. We measured the discovery strength of each cluster as its shared within-cluster edges in ProteomicsDB divided by the mean expectation of the two permutation directions (Section~\ref{sec:clusters}) and grouped the clusters into quartiles. The fraction of clusters replicated in at least one cohort rose from 38\% in the weakest quartile to 66\% in the strongest (Figure~\ref{fig:replication}d), and this trend persisted after stratifying by cluster size. In the liver cohort, which had the lowest overall replication, 69 of the 166 tested clusters fell in the weakest quartile and only 10\% of these replicated, whereas 78\% of those in the strongest quartile replicated. Many conserved co-abundance clusters therefore reflect co-abundance that is present in independent individuals rather than an artifact of the discovery data. Clusters with stronger support at discovery were also more stable in independent data.

\subsection{Biological content of conserved co-abundance clusters}
\label{sec:function}

\begin{figure}[tbp]
  \centering
  \includegraphics[width=0.8\textwidth]{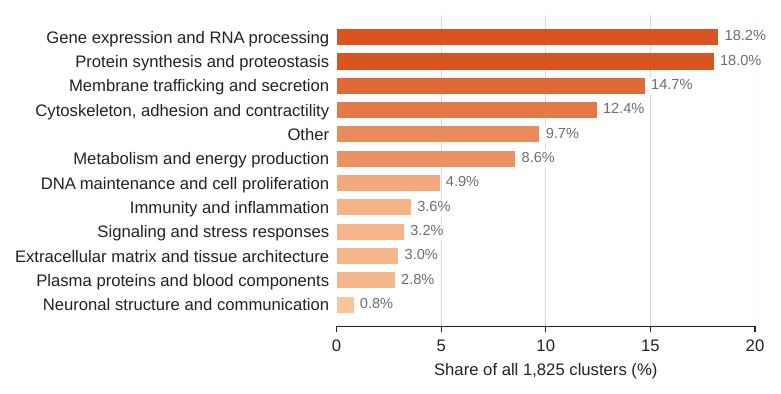}
  \caption{Functional composition of the 1{,}825 conserved co-abundance clusters. Bars show the share of all clusters in each functional category.}
  \label{fig:category-composition}
\end{figure}

Two LLM agents, Claude Code and Codex, independently assigned each conserved co-abundance cluster a principal biological theme on the basis of its member proteins and their annotations, and grouped these themes into 12 functional categories. Clusters on which the two agents disagreed were resolved by manual review (Figure~\ref{fig:category-composition}). Gene expression and RNA processing (18.2\%), protein synthesis and proteostasis (18.0\%), membrane trafficking and secretion (14.7\%), and cytoskeleton, adhesion and contractility (12.4\%) together accounted for 63\% of all clusters. Metabolism and energy production accounted for 8.6\%, and each of the remaining categories for less than 5\%.

The clusters a tissue shares reflect its physiology. DNA maintenance and cell proliferation made up 12.3\% of the co-abundance clusters shared by colon, the highest share of any tissue and well above the 4.9\% across all clusters. Cerebrospinal fluid had the highest shares of any tissue for extracellular matrix (18.6\%) and plasma proteins (17.6\%), two categories that made up only 3.0\% and 2.8\% of all clusters. Metabolism accounted for 22.5\% of the co-abundance clusters shared by heart and 17.1\% of those shared by prostate, the two highest shares among all tissues and well above the 8.6\% across all clusters.

Among these metabolic clusters, prostate and kidney shared a conserved 34-protein co-abundance cluster spanning the branched-chain amino acid degradation pathway (Table~\ref{tab:kidney-prostate}). Its members were enriched in organic acid catabolism (12 members, $q = 7.2 \times 10^{-6}$), branched-chain amino acid catabolism (7 members, $q = 2.0 \times 10^{-5}$) and valine, leucine and isoleucine degradation (8 members, $q = 1.5 \times 10^{-4}$). The cluster included BCKDHA and DBT, subunits of the branched-chain $\alpha$-keto acid dehydrogenase complex that catalyzes the irreversible oxidative decarboxylation shared by all three amino acids. It also included the downstream enzymes of each branch: IVD for leucine, ACADSB and ACAT1 for isoleucine, and ECHS1 and ALDH6A1 for valine. In both kidney and prostate, these enzymes were co-abundant, linking the shared oxidative decarboxylation step to downstream steps of all three branches. Inherited defects in these enzymes cause disorders of branched-chain amino acid metabolism, including maple syrup urine disease (BCKDHA, DBT), isovaleric acidemia (IVD) and $\beta$-ketothiolase deficiency (ACAT1), and the cluster brings these disease genes together.

Immunohistochemistry in the Human Protein Atlas detected 32 of the 34 members in kidney and 29 in prostate. In its report, the Research Agent excluded blood-derived proteins as the source of this co-abundance, as only one member is annotated as secreted and most show low immune-cell specificity. It judged the cluster to be established mitochondrial metabolism whose new aspect is its conservation between kidney and prostate. The cluster was also co-abundant in the independent CPTAC kidney cohort (Section~\ref{sec:replication}), where its 31 measured members had a mean Spearman correlation of 0.56, against a mean of 0.02 and a maximum of 0.25 for its 10{,}000 matched random sets (Figure~\ref{fig:replication}b).

\begin{table}[tbp]
  \centering
  \caption{Functional enrichment of the 34-protein co-abundance cluster shared by prostate and kidney. Genes, number of cluster members annotated to the term; Fold enrichment, their proportion in the cluster divided by the proportion of annotated proteins among all proteins in the prostate--kidney consensus network; $q$, Benjamini--Hochberg adjusted across all tests of all clusters.}
  \label{tab:kidney-prostate}
  \begin{tabular}{llccc}
    \toprule
    Source & Term & Genes & Fold enrichment & $q$ \\
    \midrule
    GO:BP & Organic acid catabolic process & 12 & 9.8 & $7.2 \times 10^{-6}$ \\
    GO:BP & Branched-chain amino acid catabolic process & 7 & 25.5 & $2.0 \times 10^{-5}$ \\
    GO:BP & Small molecule catabolic process & 13 & 7.1 & $5.0 \times 10^{-5}$ \\
    KEGG & Valine, leucine and isoleucine degradation & 8 & 14.6 & $1.5 \times 10^{-4}$ \\
    GO:BP & Amino acid catabolic process & 8 & 12.6 & $4.0 \times 10^{-4}$ \\
    \bottomrule
  \end{tabular}
\end{table}

\subsection{Conserved co-abundance clusters as a source of disease hypotheses}
\label{sec:mechanism}

In contrast to such well-characterized pathways, some conserved co-abundance clusters have few of their member relationships recorded in current interaction and complex databases and may represent coordination that has yet to be described. We examine one such cluster to illustrate how cross-tissue co-abundance can connect known protein functions, tissue context and disease evidence into a testable mechanistic hypothesis.

\begin{figure}[tbp]
  \centering
  \includegraphics[width=0.72\textwidth]{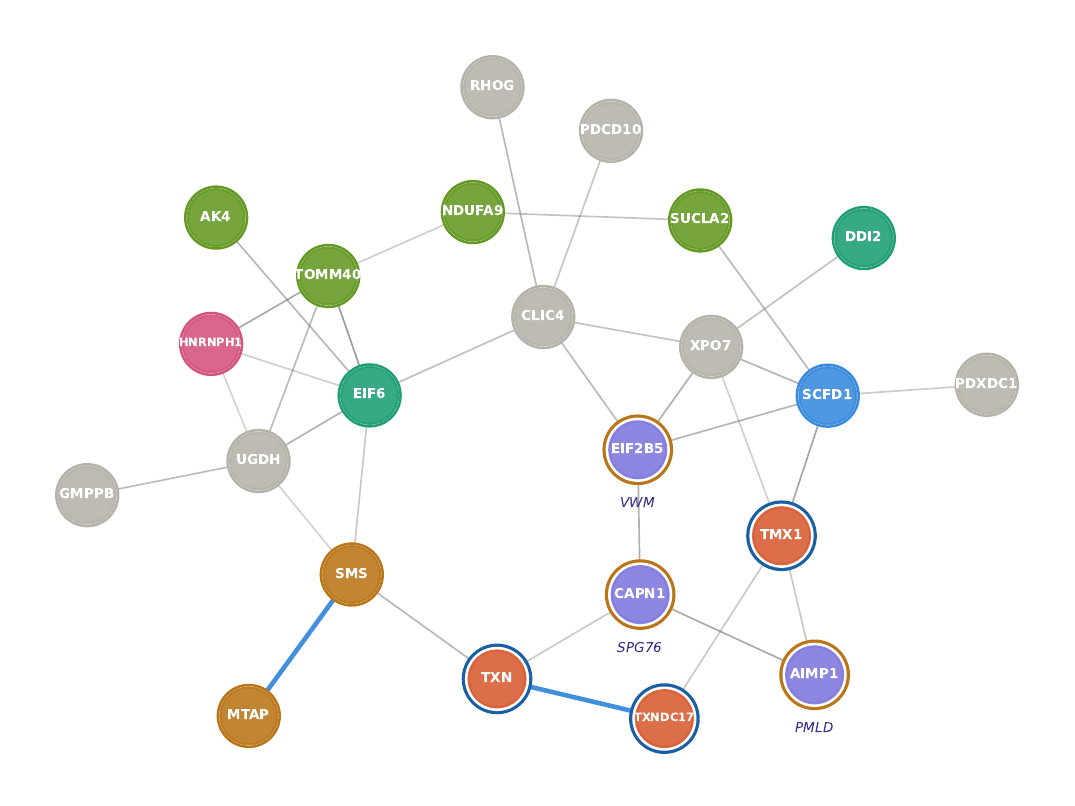}
  \caption{Co-abundance network of the conserved liver--bone marrow cluster. Nodes are colored by functional group. Gold rings mark genes linked to heritable white matter diseases (AIMP1, PMLD; CAPN1, SPG76; EIF2B5, VWM), and blue rings mark the thioredoxin redox subgroup. Blue edges are co-abundance edges that are also STRING high-confidence interactions; grey edge opacity scales with co-abundance weight.}
  \label{fig:liver-bm}
\end{figure}

Our liver--bone marrow consensus analysis recovered a conserved 23-protein co-abundance cluster (Figure~\ref{fig:liver-bm}). It includes two coherent subgroups, one involved in thioredoxin-mediated redox regulation and the other in methionine salvage and polyamine biosynthesis. GO enrichment identified protein-disulfide reductase activity as the most distinctive molecular-function signal ($q = 2.6 \times 10^{-3}$), driven by the thioredoxin-family members TXN, TXNDC17, and TMX1. STRING enrichment also highlighted this redox signature through the Redox-active center keyword (STRING FDR $= 0.011$). It also recovered a Methionine \textit{de~novo} and salvage pathway annotation (STRING FDR $= 0.0022$) linking TXN with spermine synthase (SMS) and methylthioadenosine phosphorylase (MTAP). TXN therefore provides a mechanistic bridge between these axes, consistent with its role as the electron donor for methionine sulfoxide reductases. These enzymes couple thioredoxin-dependent reducing power to the recycling of oxidized methionine residues~\cite{kim2007msr}. The co-abundance of SMS and MTAP further supports a coupled polyamine--methionine salvage pathway. SMS produces 5$'$-methylthioadenosine (MTA) during spermine biosynthesis, whereas MTAP cleaves MTA to recycle methionine (STRING combined score 986, the highest-confidence pair in the cluster).

Of the 23 members, 21 were detected in liver or bone marrow in the Human Protein Atlas. STRING records only three high-confidence interactions among the 253 possible protein pairs (TXNDC17--TXN, combined score 847; TXN--TMX1, 858; SMS--MTAP, 986), and no hu.MAP~3.0 complex covers the cluster. Current interaction and complex databases therefore do not record this 23-protein cluster as a functional unit.

The Research Agent identified a further feature of this cluster: three members, AIMP1, CAPN1 and EIF2B5, are each independently implicated in heritable white matter diseases. AIMP1 mutations cause Pelizaeus--Merzbacher-like disease~\cite{feinstein2010aimp1}, CAPN1 mutations cause hereditary spastic paraplegia type~76~\cite{ganor2016capn1}, and EIF2B5 mutations cause vanishing white matter disease (VWM)~\cite{leegwater2001eif2b}. All three are included in the 383-gene ataxia and ataxia-overlap panel compiled by the Ataxia Global Initiative~\cite{beijer2024ataxia}, yet a shared molecular basis for their convergence on white matter pathology remains unclear. Here these genes are linked by a different kind of evidence, protein co-abundance shared by liver and bone marrow. From these genes and the cluster's redox and methionine-salvage members, the Research Agent proposed a candidate mechanism. AIMP1 and EIF2B5 both participate in the GCN2--eIF2$\alpha$--eIF2B integrated stress response (ISR) axis. EIF2B5 dysfunction in VWM leads to ISR deregulation in white matter astrocytes, a phenotype pharmacologically reversible by eIF2B restoration~\cite{abbink2019}. AIMP1 encodes a component of the multi-aminoacyl-tRNA synthetase complex that connects upstream to the same axis through GCN2, a nutrient-sensing kinase whose eIF2$\alpha$ phosphorylation activity is essential for CNS remyelination~\cite{falcon2025}. The same GCN2--eIF2$\alpha$ axis is a central regulator of bone marrow hematopoietic stem cell proteostasis~\cite{li2022hsc}, offering a plausible reason why this translational checkpoint pathway surfaces as a liver--bone marrow co-abundance signal. The co-abundance of CAPN1 with AIMP1 and EIF2B5 further suggests that calpain-1 may take part in the same translational and proteostatic process whose disruption leads to white matter pathology. Co-abundance thus places three genes whose clinical convergence was recognized, but whose shared molecular logic remained unclear, in a single conserved co-abundance cluster. It points to a testable hypothesis: their white matter pathology may arise in part from dysregulation of this integrated stress response and proteostasis pathway.

\section{Conclusion and Discussion}

Cross-tissue protein coordination underlies clinically relevant disease mechanisms, yet prior proteomic studies have compared tissue pairs chosen from prior biological knowledge. Here the agents compared all 231 pairs of the available tissues under the same rules, identified 1{,}825 conserved co-abundance clusters and interpreted each against retrieved evidence. This comparison revealed relationships that targeted comparisons would miss: most tissue pairs shared co-abundance relationships, including organs with no anatomical or functional connection such as colon and lung.

Many conserved co-abundance clusters also recurred in independent CPTAC cohorts generated by other laboratories with other quantification workflows, and these clusters are therefore not specific to the resource in which they were found. Clusters with stronger support at discovery replicated more often, which makes discovery strength a practical criterion for choosing clusters to follow up.

Because conserved co-abundance clusters are enriched for known interactions, complexes and pathways, a cluster shared by two tissues points to a biological process that operates in both, rather than to similar protein levels alone. Abundance similarity reflects how close two tissues are in overall composition, whereas shared clusters reflect which processes they have in common, and the two were only weakly associated. In the cluster shared by prostate and kidney, for example, enzymes of all three branches of branched-chain amino acid degradation were co-abundant in both organs. Tissues related in this way need not be similar in composition or anatomically connected, so a mechanism studied in one tissue can be examined in the other tissues that share its cluster.

At the cluster level, the results suggest that proteins implicated in organ-confined diseases can converge within clusters shared with peripheral or anatomically distant tissues, raising the possibility that some disease programs can be studied outside their primary site of pathology, where one can test whether they also change in disease. These candidates require experimental validation, but they illustrate a category of hypothesis that targeted tissue-pair studies are unlikely to generate.

The framework does not depend on a particular set of tissues: as tissue proteomic resources expand, new tissues can be added directly and compared with every existing tissue. Co-abundance alone does not establish physical interaction, causal regulation or direct communication between tissues. The value of exhaustive, agent-assisted comparison is therefore not that it settles the biological question, but that it defines a more systematic and evidence-grounded set of questions to pursue experimentally.

\clearpage
\bibliographystyle{unsrt}
\bibliography{references}

@article{rose2025fgf21,
  title   = {{FGF21 reverses MASH through coordinated actions on the CNS and liver}},
  author  = {Rose and others},
  journal = {Cell Metab.},
  volume  = {37},
  number  = {7},
  pages   = {1515--1529},
  year    = {2025}
}

@article{kremyanskaya2024rusfertide,
  title   = {{Rusfertide, a hepcidin mimetic, for control of erythrocytosis in polycythemia vera}},
  author  = {Kremyanskaya and others},
  journal = {N. Engl. J. Med.},
  volume  = {390},
  number  = {8},
  pages   = {723--735},
  year    = {2024}
}

@article{afshar2025plasma,
  title   = {{Plasma proteomic associations with Alzheimer's disease endophenotypes}},
  author  = {Afshar and others},
  journal = {Nat. Aging},
  volume  = {5},
  number  = {10},
  pages   = {2104--2124},
  year    = {2025}
}

@article{chia2025als,
  title   = {{A plasma proteomics-based candidate biomarker panel predictive of amyotrophic lateral sclerosis}},
  author  = {Chia and others},
  journal = {Nat. Med.},
  volume  = {31},
  number  = {10},
  pages   = {3440--3450},
  year    = {2025}
}

@article{feinstein2010aimp1,
  title={Pelizaeus-Merzbacher-like disease caused by AIMP1/p43 homozygous mutation},
  author={Feinstein and others},
  journal={Am. J. Hum. Genet.},
  volume={87},
  number={6},
  pages={820--828},
  year={2010}
}

@article{ganor2016capn1,
  title={Mutations in CAPN1 cause autosomal-recessive hereditary spastic paraplegia},
  author={Gan-Or and others},
  journal={Am. J. Hum. Genet.},
  volume={98},
  number={5},
  pages={1038--1046},
  year={2016}
}

@article{leegwater2001eif2b,
  title={Subunits of the translation initiation factor eIF2B are mutant in leukoencephalopathy with vanishing white matter},
  author={Leegwater and others},
  journal={Nat. Genet.},
  volume={29},
  number={4},
  pages={383--388},
  year={2001}
}

@article{beijer2024ataxia,
  title={Standards of NGS data sharing and analysis in ataxias: recommendations by the NGS Working Group of the Ataxia Global Initiative},
  author={Beijer and others},
  journal={Cerebellum},
  volume={23},
  number={2},
  pages={391--400},
  year={2024}
}

@article{abbink2019,
  title={Vanishing white matter: deregulated integrated stress response as therapy target},
  author={Abbink and others},
  journal={Ann. Clin. Transl. Neurol.},
  volume={6},
  number={8},
  pages={1407--1422},
  year={2019}
}

@article{falcon2025,
  title={GCN2-mediated eIF2$\alpha$ phosphorylation is required for central nervous system remyelination},
  author={Falc{\'o}n and others},
  journal={Int. J. Mol. Sci.},
  volume={26},
  number={4},
  pages={1626},
  year={2025}
}

@article{li2022hsc,
  title={Amino acid catabolism regulates hematopoietic stem cell proteostasis via a GCN2-eIF2$\alpha$ axis},
  author={Li and others},
  journal={Cell Stem Cell},
  volume={29},
  number={7},
  pages={1119--1134},
  year={2022}
}

@article{kim2007msr,
  title={Methionine sulfoxide reductases: selenoprotein forms and roles in antioxidant protein repair in mammals},
  author={Kim and others},
  journal={Biochem. J.},
  volume={407},
  number={3},
  pages={321--329},
  year={2007}
}

@article{clark2019ccrcc,
  title   = {{Integrated Proteogenomic Characterization of Clear Cell Renal Cell Carcinoma}},
  author  = {Clark, David J. and others},
  journal = {Cell},
  volume  = {179},
  number  = {4},
  pages   = {964--983.e31},
  year    = {2019},
  doi     = {10.1016/j.cell.2019.10.007}
}

@article{gillette2020luad,
  title   = {{Proteogenomic Characterization Reveals Therapeutic Vulnerabilities in Lung Adenocarcinoma}},
  author  = {Gillette, Michael A. and others},
  journal = {Cell},
  volume  = {182},
  number  = {1},
  pages   = {200--225.e35},
  year    = {2020},
  doi     = {10.1016/j.cell.2020.06.013}
}

@article{satpathy2021lscc,
  title   = {{A proteogenomic portrait of lung squamous cell carcinoma}},
  author  = {Satpathy, Shankha and others},
  journal = {Cell},
  volume  = {184},
  number  = {16},
  pages   = {4348--4371.e40},
  year    = {2021},
  doi     = {10.1016/j.cell.2021.07.016}
}

@article{mcdermott2020ov,
  title   = {{Proteogenomic Characterization of Ovarian HGSC Implicates Mitotic Kinases, Replication Stress in Observed Chromosomal Instability}},
  author  = {McDermott, Jason E. and others},
  journal = {Cell Rep. Med.},
  volume  = {1},
  number  = {1},
  pages   = {100004},
  year    = {2020},
  doi     = {10.1016/j.xcrm.2020.100004}
}

@article{vasaikar2019coad,
  title   = {{Proteogenomic Analysis of Human Colon Cancer Reveals New Therapeutic Opportunities}},
  author  = {Vasaikar, Suhas and others},
  journal = {Cell},
  volume  = {177},
  number  = {4},
  pages   = {1035--1049.e19},
  year    = {2019},
  doi     = {10.1016/j.cell.2019.03.030}
}

@article{gao2019hcc,
  title   = {{Integrated Proteogenomic Characterization of HBV-Related Hepatocellular Carcinoma}},
  author  = {Gao, Qiang and others},
  journal = {Cell},
  volume  = {179},
  number  = {2},
  pages   = {561--577.e22},
  year    = {2019},
  doi     = {10.1016/j.cell.2019.08.052}
}

@article{proteomicsdb2018,
  title={ProteomicsDB},
  author={Schmidt and others},
  journal={Nucleic Acids Res.},
  volume={46},
  number={D1},
  pages={D1271--D1281},
  year={2018},
  publisher={Oxford University Press}
}

@article{uniprot2025,
  title={UniProt: the Universal Protein Knowledgebase in 2025},
  author={{The UniProt Consortium}},
  journal={Nucleic Acids Res.},
  volume={53},
  number={D1},
  pages={D609--D617},
  year={2025}
}

@article{traag2019leiden,
  title={From Louvain to Leiden: guaranteeing well-connected communities},
  author={Traag and others},
  journal={Sci. Rep.},
  volume={9},
  number={1},
  pages={5233},
  year={2019}
}

@article{benjamini1995controlling,
  title={Controlling the false discovery rate: a practical and powerful approach to multiple testing},
  author={Benjamini and others},
  journal={J. R. Stat. Soc. Ser. B Methodol.},
  volume={57},
  number={1},
  pages={289--300},
  year={1995}
}

@article{fischer2025humap3,
  title={hu.MAP3.0: atlas of human protein complexes by integration of more than 25,000 proteomic experiments},
  author={Fischer and others},
  journal={Mol. Syst. Biol.},
  volume={21},
  number={7},
  pages={911--943},
  year={2025}
}

@article{szklarczyk2023string,
  title={The STRING database in 2023: protein--protein association networks and functional enrichment analyses for any sequenced genome},
  author={Szklarczyk and others},
  journal={Nucleic Acids Res.},
  volume={51},
  number={D1},
  pages={D638--D646},
  year={2023}
}

@article{go2000,
  title={Gene ontology: tool for the unification of biology},
  author={Ashburner and others},
  journal={Nat. Genet.},
  volume={25},
  number={1},
  pages={25--29},
  year={2000}
}

@article{go2026,
  title={The Gene Ontology knowledgebase in 2026},
  author={{The Gene Ontology Consortium}},
  journal={Nucleic Acids Res.},
  volume={54},
  number={D1},
  pages={D1779--D1792},
  year={2026}
}

@article{kanehisa2000kegg,
  title={KEGG: Kyoto encyclopedia of genes and genomes},
  author={Kanehisa and others},
  journal={Nucleic Acids Res.},
  volume={28},
  number={1},
  pages={27--30},
  year={2000}
}

@article{ochoa2021opentargets,
  title={Open Targets Platform: supporting systematic drug--target identification and prioritisation},
  author={Ochoa and others},
  journal={Nucleic Acids Res.},
  volume={49},
  number={D1},
  pages={D1302--D1310},
  year={2021}
}

@article{rosonovski2024europepmc,
  title={Europe PMC in 2023},
  author={Rosonovski and others},
  journal={Nucleic Acids Res.},
  volume={52},
  number={D1},
  pages={D1668--D1676},
  year={2024}
}

@article{uhlen2015hpa,
  title   = {{Tissue-based map of the human proteome}},
  author  = {Uhl{\'e}n, Mathias and others},
  journal = {Science},
  volume  = {347},
  number  = {6220},
  pages   = {1260419},
  year    = {2015},
  doi     = {10.1126/science.1260419}
}

@article{xin2016mygene,
  title   = {{High-performance web services for querying gene and variant annotation}},
  author  = {Xin, Jiwen and others},
  journal = {Genome Biol.},
  volume  = {17},
  pages   = {91},
  year    = {2016},
  doi     = {10.1186/s13059-016-0953-9}
}

@article{pierson2015gnat,
  title   = {{Sharing and specificity of co-expression networks across 35 human tissues}},
  author  = {Pierson, Emma and Koller, Daphne and Battle, Alexis and Mostafavi, Sara and {GTEx Consortium}},
  journal = {PLoS Comput. Biol.},
  volume  = {11},
  number  = {5},
  pages   = {e1004220},
  year    = {2015},
  doi     = {10.1371/journal.pcbi.1004220}
}

@article{russell2023communities,
  title   = {{Gene communities in co-expression networks across different tissues}},
  author  = {Russell, Madison and Aqil, Alber and Saitou, Marie and Gokcumen, Omer and Masuda, Naoki},
  journal = {PLoS Comput. Biol.},
  volume  = {19},
  number  = {11},
  pages   = {e1011616},
  year    = {2023},
  doi     = {10.1371/journal.pcbi.1011616}
}

@article{kustatscher2019proteomehd,
  title   = {{Co-regulation map of the human proteome enables identification of protein functions}},
  author  = {Kustatscher, Georg and Grabowski, Piotr and Schrader, Tina A. and Passmore, Josiah B. and Schrader, Michael and Rappsilber, Juri},
  journal = {Nat. Biotechnol.},
  volume  = {37},
  pages   = {1361--1371},
  year    = {2019},
  doi     = {10.1038/s41587-019-0298-5}
}


\end{document}